\documentclass[sigconf]{acmart}

\usepackage{booktabs} 
\usepackage{multirow}

\setcopyright{rightsretained}

\acmDOI{10.475/123_4}

\acmISBN{123-4567-24-567/08/06}

\acmConference[SIDEWAYS'17]{Workshop on Social Media World Sensors (SIDEWAYS)}{July 2017}{Prague, Czech Republic}
\acmYear{2017}
\copyrightyear{2017}

\acmPrice{15.00}

\begin{document}
\title{Stance Detection in Turkish Tweets}

\author{Dilek K\"u\c{c}\"uk}
\orcid{0000-0003-2656-1300}
\affiliation{
  \institution{Electrical Power Technologies Group\\T\"UB\.ITAK Energy Institute}
  \city{Ankara}
  \state{Turkey}
}
\email{dilek.kucuk@tubitak.gov.tr}

\renewcommand{\shortauthors}{D. K\"u\c{c}\"uk}

\begin{abstract}
Stance detection is a classification problem in natural language processing where for a text and target pair, a class result from the set \{\emph{Favor}, \emph{Against}, \emph{Neither}\} is expected. It is similar to the sentiment analysis problem but instead of the sentiment of the text author, the stance expressed for a particular target is investigated in stance detection. In this paper, we present a stance detection tweet data set for Turkish comprising stance annotations of these tweets for two popular sports clubs as targets. Additionally, we provide the evaluation results of SVM classifiers for each target on this data set, where the classifiers use unigram, bigram, and hashtag features. This study is significant as it presents one of the initial stance detection data sets proposed so far and the first one for Turkish language, to the best of our knowledge. The data set and the evaluation results of the corresponding SVM-based approaches will form plausible baselines for the comparison of future studies on stance detection.
\end{abstract}




\maketitle

\section{Introduction}\label{intro}
Stance detection (also called \emph{stance identification} or \emph{stance classification}) is one of the considerably recent research topics in natural language processing (NLP). It is usually defined as a classification problem where for a text and target pair, the stance of the author of the text for that target is expected as a classification output from the set: \{\emph{Favor}, \emph{Against}, \emph{Neither}\} \cite{mohammad2016semeval}.

Stance detection is usually considered as a subtask of sentiment analysis (opinion mining) \cite{pang2008opinion} topic in NLP. Both are mostly performed on social media texts, particularly on tweets, hence both are important components of social media analysis. Nevertheless, in sentiment analysis, the sentiment of the author of a piece of text usually as \emph{Positive}, \emph{Negative}, and \emph{Neutral} is explored while in stance detection, the stance of the author of the text for a particular target (an entity, event, etc.) either explicitly or implicitly referred to in the text is considered. Like sentiment analysis, stance detection systems can be valuable components of information retrieval and other text analysis systems \cite{mohammad2016semeval}.

Previous work on stance detection include \cite{somasundaran2010recognizing} where a stance classifier based on sentiment and arguing features is proposed in addition to an arguing lexicon automatically compiled. The ultimate approach performs better than distribution-based and uni-gram-based baseline systems \cite{somasundaran2010recognizing}. In \cite{walker2012stance}, the authors show that the use of dialogue structure improves stance detection in on-line debates. In \cite{hasan2013stance}, Hasan and Ng carry out stance detection experiments using different machine learning algorithms, training data sets, features, and inter-post constraints in on-line debates, and draw insightful conclusions based on these experiments. For instance, they find that sequence models like HMMs perform better at stance detection when compared with non-sequence models like Naive Bayes (NB) \cite{hasan2013stance}. In another related study \cite{misra2013topic}, the authors conclude that topic-independent features can be exploited for disagreement detection in on-line dialogues. The employed features include agreement, cue words, denial, hedges, duration, polarity, and punctuation \cite{misra2013topic}. Stance detection on a corpus of student essays is considered in \cite{faulkner2014automated}. After using linguistically-motivated feature sets together with multivalued NB and SVM as the learning models, the authors conclude that they outperform two baseline approaches \cite{faulkner2014automated}. In \cite{dori2015controversy}, the author claims that Wikipedia can be used to determine stances about controversial topics based on their previous work regarding controversy extraction on the Web.

Among more recent related work, in \cite{augenstein2016stance} stance detection for unseen targets is studied and bidirectional conditional encoding is employed. The authors state that their approach achieves state-of-the art performance rates \cite{augenstein2016stance} on \emph{SemEval 2016 Twitter Stance Detection} corpus \cite{mohammad2016semeval}. In \cite{chen2016scifnet}, a stance-community detection approach called SCIFNET is proposed. SCIFNET creates networks of people who are stance targets, automatically from the related document collections \cite{chen2016scifnet} using stance expansion and refinement techniques to arrive at stance-coherent networks. A tweet data set annotated with stance information regarding six predefined targets is proposed in \cite{mohammad2016dataset} where this data set is annotated through crowdsourcing. The authors indicate that the data set is also annotated with sentiment information in addition to stance, so it can help reveal associations between stance and sentiment \cite{mohammad2016dataset}. Lastly, in \cite{mohammad2016semeval}, SemEval 2016's aforementioned shared task on \emph{Twitter Stance Detection} is described. Also provided are the results of the evaluations of 19 systems participating in two subtasks (one with training data set provided and the other without an annotated data set) of the shared task \cite{mohammad2016semeval}.

In this paper, we present a tweet data set in Turkish annotated with stance information, where the corresponding annotations are made publicly available. The domain of the tweets comprises two popular football clubs which constitute the targets of the tweets included. We also provide the evaluation results of SVM classifiers (for each target) on this data set using unigram, bigram, and hashtag features.

To the best of our knowledge, the current study is the first one to target at stance detection in Turkish tweets. Together with the provided annotated data set and the corresponding evaluations with the aforementioned SVM classifiers which can be used as baseline systems, our study will hopefully help increase social media analysis studies on Turkish content.

The rest of the paper is organized as follows: In Section \ref{dataset}, we describe our tweet data set annotated with the target and stance information. Section \ref{svm} includes the details of our SVM-based stance classifiers and their evaluation results with discussions. Section \ref{future} includes future research topics based on the current study, and finally Section \ref{conc} concludes the paper with a summary.

\section{A Stance Detection Data Set}\label{dataset}
We have decided to consider tweets about popular sports clubs as our domain for stance detection. Considerable amounts of tweets are being published for sports-related events at every instant. Hence we have determined our targets as \emph{Galatasaray} (namely \emph{Target-1}) and \emph{Fenerbah\c{c}e} (namely, \emph{Target-2}) which are two of the most popular football clubs in Turkey. As is the case for the sentiment analysis tools, the outputs of the stance detection systems on a stream of tweets about these clubs can facilitate the use of the opinions of the football followers by these clubs.

In a previous study on the identification of public health-related tweets, two tweet data sets in Turkish (each set containing 1 million random tweets) have been compiled where these sets belong to two different periods of 20 consecutive days \cite{Kucuk2017}. We have decided to use one of these sets (corresponding to the period between August 18 and September 6, 2015) and firstly filtered the tweets using the possible names used to refer to the target clubs. Then, we have annotated the stance information in the tweets for these targets as \emph{Favor} or \emph{Against}. Within the course of this study, we have not considered those tweets in which the target is not explicitly mentioned, as our initial filtering process reveals.

For the purposes of the current study, we have not annotated any tweets with the \emph{Neither} class. This stance class and even finer-grained classes can be considered in further annotation studies. We should also note that in a few tweets, the target of the stance was the management of the club while in some others a particular footballer of the club is praised or criticised. Still, we have considered the club as the target of the stance in all of the cases and carried out our annotations accordingly.

At the end of the annotation process, we have annotated 700 tweets, where 175 tweets are in favor of and 175 tweets are against \emph{Target-1}, and similarly 175 tweets are in favor of and 175 are against \emph{Target-2}. Hence, our data set is a balanced one although it is currently limited in size. The corresponding stance annotations are made publicly available at \texttt{http://ceng.metu.edu.tr/}$\sim$\texttt{e120329/} \texttt{Turkish\_Stance\_Detection\_Tweet\_Dataset.csv} in Comma Separated Values (CSV) format. The file contains three columns with the corresponding headers. The first column is the tweet id of the corresponding tweet, the second column contains the name of the stance target, and the last column includes the stance of the tweet for the target as \emph{Favor} or \emph{Against}.

To the best of our knowledge, this is the first publicly-available stance-annotated data set for Turkish. Hence, it is a significant resource as there is a scarcity of annotated data sets, linguistic resources, and NLP tools available for Turkish. Additionally, to the best of our knowledge, it is also significant for being the first stance-annotated data set including sports-related tweets, as previous stance detection data sets mostly include on-line texts on political\slash ethical issues.

\section{Stance Detection Experiments Using SVM Classifiers}\label{svm}
It is emphasized in the related literature that unigram-based methods are reliable for the stance detection task \cite{somasundaran2010recognizing} and similarly unigram-based models have been used as baseline models in studies such as \cite{mohammad2016semeval}. In order to be used as a baseline and reference system for further studies on stance detection in Turkish tweets, we have trained two SVM classifiers (one for each target) using unigrams as features. Before the extraction of unigrams, we have employed automated preprocessing to filter out the stopwords in our annotated data set of 700 tweets. The stopword list used is the list presented in \cite{kucuk2011exploiting} which, in turn, is the slightly extended version of the stopword list provided in \cite{can2008information}.

We have used the SVM implementation available in the Weka data mining application \cite{weka2009} where this particular implementation employs the SMO algorithm \cite{platt1999} to train a classifier with a linear kernel. The 10-fold cross-validation results of the two classifiers are provided in Table \ref{tab:result1} using the metrics of precision, recall, and F-Measure.

\begin{table}[h!]
\centering
\caption{Evaluation Results of the Unigram-based SVM Classifiers}
\label{tab:result1}
\begin{tabular}{|l|l|c|c|c|}
\hline
Target & Class & Precision (\%) & Recall (\%) & F-Measure (\%) \\
\hline
\multirow{3}{*}{Target-1} & Favor & 75.2 & 92.0 & 82.8 \\
                            & Against & 89.7 & 69.7 & 78.5 \\
                             & \textbf{Average} & \textbf{82.5} & \textbf{80.9} & \textbf{80.6} \\
                             \hline
\multirow{3}{*}{Target-2} & Favor & 68.5 & 83.4 & 75.3 \\
                            & Against & 78.8 & 61.7 & 69.2 \\
                             & \textbf{Average} & \textbf{73.7} & \textbf{72.6} & \textbf{72.2} \\
\hline
\end{tabular}
\end{table}

The evaluation results are quite favorable for both targets and particularly higher for \emph{Target-1}, considering the fact that they are the initial experiments on the data set. The performance of the classifiers is better for the \emph{Favor} class for both targets when compared with the performance results for the \emph{Against} class. This outcome may be due to the common use of some terms when expressing positive stance towards sports clubs in Turkish tweets. The same percentage of common terms may not have been observed in tweets during the expression of negative stances towards the targets. Yet, completely the opposite pattern is observed in stance detection results of baseline systems given in \cite{mohammad2016semeval}, i.e., better F-Measure rates have been obtained for the \emph{Against} class when compared with the \emph{Favor} class \cite{mohammad2016semeval}. Some of the baseline systems reported in \cite{mohammad2016semeval} are SVM-based systems using unigrams and ngrams as features similar to our study, but their data sets include all three stance classes of \emph{Favor}, \emph{Against}, and \emph{Neither}, while our data set comprises only tweets classified as belonging to \emph{Favor} or \emph{Against} classes. Another difference is that the data sets in \cite{mohammad2016semeval} have been divided into training and test sets, while in our study we provide 10-fold cross-validation results on the whole data set. On the other hand, we should also note that SVM-based sentiment analysis systems (such as those given in \cite{poursepanj2013uottawa}) have been reported to achieve better F-Measure rates for the \emph{Positive} sentiment class when compared with the results obtained for the \emph{Negative} class. Therefore, our evaluation results for each stance class seem to be in line with such sentiment analysis systems. Yet, further experiments on the extended versions of our data set should be conducted and the results should again be compared with the stance detection results given in the literature.

We have also evaluated SVM classifiers which use only bigrams as features, as ngram-based classifiers have been reported to perform better for the stance detection problem \cite{mohammad2016semeval}. However, we have observed that using bigrams as the sole features of the SVM classifiers leads to quite poor results. This observation may be due to the relatively limited size of the tweet data set employed. Still, we can conclude that unigram-based features lead to superior results compared to the results obtained using bigrams as features, based on our experiments on our data set. Yet, ngram-based features may be employed on the extended versions of the data set to verify this conclusion within the course of future work.

With an intention to exploit the contribution of hashtag use to stance detection, we have also used the existence of hashtags in tweets as an additional feature to unigrams. The corresponding evaluation results of the SVM classifiers using unigrams together the existence of hashtags as features are provided in Table \ref{tab:result2}.

\begin{table}[h!]
\centering
\caption{Evaluation Results of the SVM Classifiers Utilizing Unigrams and Hashtag Use as Features}
\label{tab:result2}
\begin{tabular}{|l|l|c|c|c|}
\hline
Target & Class & Precision (\%) & Recall (\%) & F-Measure (\%) \\
\hline
\multirow{3}{*}{Target-1} & Favor & 75.0 & 90.9 & 82.2 \\
                            & Against & 88.4 & 69.7 & 78.0 \\
                             & \textbf{Average} & \textbf{81.7} & \textbf{80.3} & \textbf{80.1} \\
                             \hline
\multirow{3}{*}{Target-2} & Favor & 70.0 & 85.1 & 76.8 \\
                            & Against & 81.0 & 63.4 & 71.2 \\
                             & \textbf{Average} & \textbf{75.5} & \textbf{74.3} & \textbf{74.0} \\
\hline
\end{tabular}
\end{table}

When the results given in Table \ref{tab:result2} are compared with the results in Table \ref{tab:result1}, a slight decrease in F-Measure (0.5\%) for \emph{Target-1} is observed, while the overall F-Measure value for \emph{Target-2} has increased by 1.8\%. Although we could not derive sound conclusions mainly due to the relatively small size of our data set, the increase in the performance of the SVM classifier \emph{Target-2} is an encouraging evidence for the exploitation of hashtags in a stance detection system. We leave other ways of exploiting hashtags for stance detection as a future work.

To sum up, our evaluation results are significant as reference results to be used for comparison purposes and provides evidence for the utility of unigram-based and hashtag-related features in SVM classifiers for the stance detection problem in Turkish tweets.

\section{Future Prospects}\label{future}

Future work based on the current study includes the following:
\begin{itemize}
  \item The presented stance-annotated data set for Turkish has been created by one annotator only (the author of this study), yet, the data set should better be revised and extended through crowdsourcing facilities. When employing such a procedure, other stance classes like \emph{Neither} can be considered as well. The procedure will improve the quality the data set as well as the quality of prospective systems to be trained and tested on it.
  \item Other features like emoticons (as commonly used for sentiment analysis), features based on hashtags, and ngram features can also be used by the classifiers and these classifiers can be tested on larger data sets. Other classification approaches could also be implemented and tested against our baseline classifiers. Particularly, related methods presented in recent studies such as \cite{mohammad2016semeval} can be tested on our data set.
  \item Lastly, the SVM classifiers utilized in this study and their prospective versions utilizing other features can be tested on stance data sets in other languages (such as English) for comparison purposes.
\end{itemize}

\section{Conclusion}\label{conc}
Stance detection is a considerably new research area in natural language processing and is considered within the scope of the well-studied topic of sentiment analysis. It is the detection of stance within text towards a target which may be explicitly specified in the text or not. In this study, we present a stance-annotated tweet data set in Turkish where the targets of the annotated stances are two popular sports clubs in Turkey. The corresponding annotations are made publicly-available for research purposes. To the best of our knowledge, this is the first stance detection data set for the Turkish language and also the first sports-related stance-annotated data set. Also presented in this study are SVM classifiers (one for each target) utilizing unigram and bigram features in addition to using the existence of hashtags as another feature. 10-fold cross validation results of these classifiers are presented which can be used as reference results by prospective systems. Both the annotated data set and the classifiers with evaluations are significant since they are the initial contributions to stance detection problem in Turkish tweets.

\bibliographystyle{ACM-Reference-Format}

\begin{thebibliography}{00}


\ifx \showCODEN    \undefined \def \showCODEN     #1{\unskip}     \fi
\ifx \showDOI      \undefined \def \showDOI       #1{{\tt DOI:}\penalty0{#1}\ }
  \fi
\ifx \showISBNx    \undefined \def \showISBNx     #1{\unskip}     \fi
\ifx \showISBNxiii \undefined \def \showISBNxiii  #1{\unskip}     \fi
\ifx \showISSN     \undefined \def \showISSN      #1{\unskip}     \fi
\ifx \showLCCN     \undefined \def \showLCCN      #1{\unskip}     \fi
\ifx \shownote     \undefined \def \shownote      #1{#1}          \fi
\ifx \showarticletitle \undefined \def \showarticletitle #1{#1}   \fi
\ifx \showURL      \undefined \def \showURL       #1{#1}          \fi
\providecommand\bibfield[2]{#2}
\providecommand\bibinfo[2]{#2}
\providecommand\natexlab[1]{#1}
\providecommand\showeprint[2][]{arXiv:#2}

\bibitem[\protect\citeauthoryear{Augenstein, Rockt{\"a}schel, Vlachos, and
  Bontcheva}{Augenstein et~al\mbox{.}}{2016}]%
        {augenstein2016stance}
\bibfield{author}{\bibinfo{person}{Isabelle Augenstein}, \bibinfo{person}{Tim
  Rockt{\"a}schel}, \bibinfo{person}{Andreas Vlachos}, {and}
  \bibinfo{person}{Kalina Bontcheva}.} \bibinfo{year}{2016}\natexlab{}.
\newblock \showarticletitle{Stance detection with bidirectional conditional
  encoding}. In \bibinfo{booktitle}{{\em Proceedings of the Conference on
  Empirical Methods in Natural Language Processing (EMNLP)}}.
\newblock


\bibitem[\protect\citeauthoryear{Can, Kocberber, Balcik, Kaynak, Ocalan, and
  Vursavas}{Can et~al\mbox{.}}{2008}]%
        {can2008information}
\bibfield{author}{\bibinfo{person}{Fazli Can}, \bibinfo{person}{Seyit
  Kocberber}, \bibinfo{person}{Erman Balcik}, \bibinfo{person}{Cihan Kaynak},
  \bibinfo{person}{H~Cagdas Ocalan}, {and} \bibinfo{person}{Onur~M Vursavas}.}
  \bibinfo{year}{2008}\natexlab{}.
\newblock \showarticletitle{Information retrieval on Turkish texts}.
\newblock \bibinfo{journal}{{\em Journal of the American Society for
  Information Science and Technology\/}} \bibinfo{volume}{59},
  \bibinfo{number}{3} (\bibinfo{year}{2008}), \bibinfo{pages}{407--421}.
\newblock


\bibitem[\protect\citeauthoryear{Chen and Chen}{Chen and Chen}{2016}]%
        {chen2016scifnet}
\bibfield{author}{\bibinfo{person}{Zhong-Yong Chen} {and}
  \bibinfo{person}{Chien~Chin Chen}.} \bibinfo{year}{2016}\natexlab{}.
\newblock \showarticletitle{SCIFNET: Stance community identification of topic
  persons using friendship network analysis}.
\newblock \bibinfo{journal}{{\em Knowledge-Based Systems\/}}
  \bibinfo{volume}{110} (\bibinfo{year}{2016}), \bibinfo{pages}{30--48}.
\newblock


\bibitem[\protect\citeauthoryear{Dori-Hacohen}{Dori-Hacohen}{2015}]%
        {dori2015controversy}
\bibfield{author}{\bibinfo{person}{Shiri Dori-Hacohen}.}
  \bibinfo{year}{2015}\natexlab{}.
\newblock \showarticletitle{Controversy Detection and Stance Analysis}. In
  \bibinfo{booktitle}{{\em Proceedings of the 38th International ACM SIGIR
  Conference on Research and Development in Information Retrieval}}.
  \bibinfo{pages}{1057--1057}.
\newblock


\bibitem[\protect\citeauthoryear{Faulkner}{Faulkner}{2014}]%
        {faulkner2014automated}
\bibfield{author}{\bibinfo{person}{Adam Faulkner}.}
  \bibinfo{year}{2014}\natexlab{}.
\newblock \showarticletitle{Automated classification of stance in student
  essays: An approach using stance target information and the Wikipedia
  link-based measure}. In \bibinfo{booktitle}{{\em Proceedings of the
  Twenty-Seventh International Florida Artificial Intelligence Research Society
  Conference}}.
\newblock


\bibitem[\protect\citeauthoryear{Hall, Frank, Holmes, Pfahringer, Reutemann,
  and Witten}{Hall et~al\mbox{.}}{2009}]%
        {weka2009}
\bibfield{author}{\bibinfo{person}{M. Hall}, \bibinfo{person}{E. Frank},
  \bibinfo{person}{G. Holmes}, \bibinfo{person}{B. Pfahringer},
  \bibinfo{person}{P. Reutemann}, {and} \bibinfo{person}{I.~H. Witten}.}
  \bibinfo{year}{2009}\natexlab{}.
\newblock \showarticletitle{The WEKA Data Mining Software: An Update}.
\newblock \bibinfo{journal}{{\em ACM SIGKDD Explorations Newsletter\/}}
  \bibinfo{volume}{1}, \bibinfo{number}{1} (\bibinfo{year}{2009}),
  \bibinfo{pages}{10--18}.
\newblock


\bibitem[\protect\citeauthoryear{Hasan and Ng}{Hasan and Ng}{2013}]%
        {hasan2013stance}
\bibfield{author}{\bibinfo{person}{Kazi~Saidul Hasan} {and}
  \bibinfo{person}{Vincent Ng}.} \bibinfo{year}{2013}\natexlab{}.
\newblock \showarticletitle{Stance Classification of Ideological Debates: Data,
  Models, Features, and Constraints}. In \bibinfo{booktitle}{{\em Proceedings
  of the Sixth International Joint Conference on Natural Language Processing}}.
  \bibinfo{pages}{1348--1356}.
\newblock


\bibitem[\protect\citeauthoryear{K\"u\c{c}\"uk}{K\"u\c{c}\"uk}{2011}]%
        {kucuk2011exploiting}
\bibfield{author}{\bibinfo{person}{Dilek K\"u\c{c}\"uk}.}
  \bibinfo{year}{2011}\natexlab{}.
\newblock {\em \bibinfo{title}{Exploiting Information Extraction Techniques for
  Automatic Semantic Annotation and Retrieval of News Videos in Turkish}}.
\newblock \bibinfo{thesistype}{Ph.D. Dissertation}. \bibinfo{school}{Middle
  East Technical University}.
\newblock


\bibitem[\protect\citeauthoryear{K\"u\c{c}\"uk, Yapar, K\"u\c{c}\"uk, and
  K\"u\c{c}\"uk}{K\"u\c{c}\"uk et~al\mbox{.}}{2017}]%
        {Kucuk2017}
\bibfield{author}{\bibinfo{person}{Emine~Ela K\"u\c{c}\"uk},
  \bibinfo{person}{K\"ur\c{s}ad Yapar}, \bibinfo{person}{Dilek K\"u\c{c}\"uk},
  {and} \bibinfo{person}{Do\u{g}an K\"u\c{c}\"uk}.}
  \bibinfo{year}{2017}\natexlab{}.
\newblock \showarticletitle{Ontology-based automatic identification of public
  health-related {Turkish} tweets}.
\newblock \bibinfo{journal}{{\em Computers in Biology and Medicine\/}}
  \bibinfo{volume}{83} (\bibinfo{year}{2017}), \bibinfo{pages}{1--9}.
\newblock


\bibitem[\protect\citeauthoryear{Misra and Walker}{Misra and Walker}{2013}]%
        {misra2013topic}
\bibfield{author}{\bibinfo{person}{Amita Misra} {and}
  \bibinfo{person}{Marilyn~A Walker}.} \bibinfo{year}{2013}\natexlab{}.
\newblock \showarticletitle{Topic independent identification of agreement and
  disagreement in social media dialogue}. In \bibinfo{booktitle}{{\em
  Conference of the Special Interest Group on Discourse and Dialogue}}.
  \bibinfo{pages}{920}.
\newblock


\bibitem[\protect\citeauthoryear{Mohammad, Kiritchenko, Sobhani, Zhu, and
  Cherry}{Mohammad et~al\mbox{.}}{2016a}]%
        {mohammad2016dataset}
\bibfield{author}{\bibinfo{person}{Saif~M Mohammad}, \bibinfo{person}{Svetlana
  Kiritchenko}, \bibinfo{person}{Parinaz Sobhani}, \bibinfo{person}{Xiaodan
  Zhu}, {and} \bibinfo{person}{Colin Cherry}.}
  \bibinfo{year}{2016}\natexlab{a}.
\newblock \showarticletitle{A dataset for detecting stance in tweets}. In
  \bibinfo{booktitle}{{\em Proceedings of the Language Resources and Evaluation
  Conference (LREC)}}.
\newblock


\bibitem[\protect\citeauthoryear{Mohammad, Kiritchenko, Sobhani, Zhu, and
  Cherry}{Mohammad et~al\mbox{.}}{2016b}]%
        {mohammad2016semeval}
\bibfield{author}{\bibinfo{person}{Saif~M Mohammad}, \bibinfo{person}{Svetlana
  Kiritchenko}, \bibinfo{person}{Parinaz Sobhani}, \bibinfo{person}{Xiaodan
  Zhu}, {and} \bibinfo{person}{Colin Cherry}.}
  \bibinfo{year}{2016}\natexlab{b}.
\newblock \showarticletitle{Semeval-2016 task 6: Detecting stance in tweets}.
  In \bibinfo{booktitle}{{\em Proceedings of the International Workshop on
  Semantic Evaluation, SemEval}}.
\newblock


\bibitem[\protect\citeauthoryear{Pang and Lee}{Pang and Lee}{2008}]%
        {pang2008opinion}
\bibfield{author}{\bibinfo{person}{Bo Pang} {and} \bibinfo{person}{Lillian
  Lee}.} \bibinfo{year}{2008}\natexlab{}.
\newblock \showarticletitle{Opinion mining and sentiment analysis}.
\newblock \bibinfo{journal}{{\em Foundations and Trends in Information
  Retrieval\/}} \bibinfo{volume}{2}, \bibinfo{number}{1-2}
  (\bibinfo{year}{2008}), \bibinfo{pages}{1--135}.
\newblock


\bibitem[\protect\citeauthoryear{Platt}{Platt}{1999}]%
        {platt1999}
\bibfield{author}{\bibinfo{person}{John~C. Platt}.}
  \bibinfo{year}{1999}\natexlab{}.
\newblock \showarticletitle{Fast Training of Support Vector Machines Using
  Sequential Minimal Optimization}.
\newblock \bibinfo{journal}{{\em Advances in Kernel Methods\/}}
  (\bibinfo{year}{1999}), \bibinfo{pages}{185--208}.
\newblock


\bibitem[\protect\citeauthoryear{Poursepanj, Weissbock, and Inkpen}{Poursepanj
  et~al\mbox{.}}{2013}]%
        {poursepanj2013uottawa}
\bibfield{author}{\bibinfo{person}{Hamid Poursepanj}, \bibinfo{person}{Josh
  Weissbock}, {and} \bibinfo{person}{Diana Inkpen}.}
  \bibinfo{year}{2013}\natexlab{}.
\newblock \showarticletitle{uOttawa: System description for SemEval 2013 Task 2
  Sentiment Analysis in Twitter}. In \bibinfo{booktitle}{{\em Second Joint
  Conference on Lexical and Computational Semantics (*SEM), Volume 2: Seventh
  International Workshop on Semantic Evaluation (SemEval 2013)}}.
  \bibinfo{pages}{380--383}.
\newblock


\bibitem[\protect\citeauthoryear{Somasundaran and Wiebe}{Somasundaran and
  Wiebe}{2010}]%
        {somasundaran2010recognizing}
\bibfield{author}{\bibinfo{person}{Swapna Somasundaran} {and}
  \bibinfo{person}{Janyce Wiebe}.} \bibinfo{year}{2010}\natexlab{}.
\newblock \showarticletitle{Recognizing stances in ideological on-line
  debates}. In \bibinfo{booktitle}{{\em Proceedings of the NAACL HLT Workshop
  on Computational Approaches to Analysis and Generation of Emotion in Text}}.
  \bibinfo{pages}{116--124}.
\newblock


\bibitem[\protect\citeauthoryear{Walker, Anand, Abbott, and Grant}{Walker
  et~al\mbox{.}}{2012}]%
        {walker2012stance}
\bibfield{author}{\bibinfo{person}{Marilyn~A Walker}, \bibinfo{person}{Pranav
  Anand}, \bibinfo{person}{Robert Abbott}, {and} \bibinfo{person}{Ricky
  Grant}.} \bibinfo{year}{2012}\natexlab{}.
\newblock \showarticletitle{Stance classification using dialogic properties of
  persuasion}. In \bibinfo{booktitle}{{\em Proceedings of the Conference of the
  North American Chapter of the Association for Computational Linguistics:
  Human Language Technologies}}. \bibinfo{pages}{592--596}.
\newblock


\end{thebibliography}

\end{document}